# Deep Learning-Based Approach for Identification of Potato Leaf Diseases Using Wrapper Feature Selection and Feature Concatenation

[1]Muhammad Ahtsam Naeem, [2]Muhammad Asim Saleem, [3]Muhammad Imran Sharif, [2]Shahzad Akber, [4]Sajjad Saleem, [5]Zahid Akhtar and [6]Kamran Siddique

[1]*School of Information and Software Engineering, University of Electronic Science and Technology of China, China*
[2]*Faculty of Computing, Riphah International University, Faisalabad Campus, Pakistan.*
[3]*Department of Computer Science, COMSATS University Islamabad, Wah Campus, Pakistan*
[4]*Department of Information and Technology, Washington University of Science and Technology, Alexandria, VA, USA*
[5]*Department of Network and Computer Security, State University of New York Polytechnic Institute, USA*
[6]*Department of Computer Science and Engineering, University of Alaska Anchorage, Anchorage, USA*

Corresponding Author:
Sajjad Saleem
Department of Information and Technology, Washington University of Science and Technology, Alexandria, VA, USA
Email: ssaleem.student@wust.edu

**Abstract:** The potato is a widely grown crop in many regions of the world. In recent decades, potato farming has gained incredible traction in the world. Potatoes are susceptible to several illnesses that stunt their development. This plant seems to have significant leaf disease. Early Blight and Late Blight are two prevalent leaf diseases that affect potato plants. The early detection of these diseases would be beneficial for enhancing the yield of this crop. The ideal solution is to use image processing to identify and analyze these disorders. Here, we present an autonomous method based on image processing and machine learning to detect late blight disease affecting potato leaves. The proposed method comprises four different phases: (1) Histogram Equalization is used to improve the quality of the input image; (2) Feature extraction is performed using a Deep CNN model, then these extracted features are concatenated; (3) Feature selection is performed using wrapper-based feature selection; (4) Classification is performed using an SVM classifier and its variants. This proposed method achieves the highest accuracy of 99% using SVM by selecting 550 features.

## Introduction

In areas where potatoes are planted, late blight diseases are frequent. According to Arora *et al.*, potato blight is a common green leaf disease that begins as irregular light green lesions near the tip and margins of the leaf and then spreads into large brown to purplish-black necrotic patches (Arora *et al.*, 2014; Arnaud *et al.*, 2022). Its production of over 329 million metric tons of non-grain agricultural products in 2009 serves as an essential element of over 1.5 billion people's daily diets (Muthoni & Nyamongo, 2009). Potato is a multipurpose and widely available crop that has played a critical role in China's economic growth. Pests and diseases, on the other hand, significantly limit potato production. Late blight, a disease that affects potatoes all over the globe, is the most destructive (Ugonna *et al*., 2013). Farmers who want to minimize annual losses must first identify the many illnesses that harm potato plants. Early blight and late blight are two of the most frequent diseases. Economic loss and waste reduction prevention may be achieved by early identification of these illnesses and the subsequent application of effective therapy. Because the treatments for early blight and late blight are distinct, every potato plant must be correctly identified. In this study, researchers use deep learning to develop a model that uses convolutional neural networks to classify these potato plant diseases. For farmers to select the proper treatment, the model should provide a way for them to identify whether their potato plants have early or late blight diseases (El Moutaouakil *et al*., 2022).

Environmental factors have an impact on the occurrence of late blight. Excessive humidity and specific temperature conditions may wreak havoc on an entire field in a short period (Hou *et al*., 2022). Leaf yellowing, wilting, and withering will be seen, as well as a shift in spectral characteristics (Mora-Romero *et al*., 2022). Islam *et al*. (2017) suggested a method for diagnosing plant leaf disease that includes image processing and machine learning. An automated technique called 'Plant Village' has been used to identify potato plant illnesses and natural leaves from the public image. The segmentation and Classification described here are carried out using support vector machines, and the suggested model's accuracy is around 95%. A large-scale plant disease diagnosis may now be made using the suggested method. The multiclass SVM image segmentation was applied to build an automated and user-friendly system. A small amount of computational work is required to identify basic potato illnesses such as Late Blight and Early Blight. Using this strategy, farmers could identify diseases more quickly, accurately, and efficiently (Islam *et al*., 2017).

Agriculture nowadays is complex (Muiz Fayyaz *et al*., 2022). Regarding farming, the business has evolved into an increasingly competitive and worldwide one, where producers have to consider local climatic conditions, environmental issues, and worldwide environmental and economic elements (Tsiafouli *et al*., 2017). Compared to grains, the potato is a more productive crop in terms of the amount of protein, dry matter, and minerals it generates per unit of land. On the other hand, the production of potatoes is susceptible to a wide variety of illnesses, which may cause yield losses and a decline in tuber quality, ultimately leading to an increase in the price of potatoes (Saleem *et al*., 2024). Potato crops are susceptible to a wide variety of illnesses, particularly parasitic infections, which may result in major decreases in crop output and severe financial losses for farmers and producers (Sjölin *et al*., 2022; Mushtaq *et al*., 2022).

The agriculture industry plays a vital role in the world economy (Muiz Fayyaz *et al*., 2022). It emphasizes the need to provide adequate care for plants from the time they are seedlings until they produce the desired crop. The crop has to undergo several stages throughout this procedure (Mrożek and Koronowicz, (n.d.)). This problem may be resolved if enough safeguards are put in place to secure the playing area. There are some aspects of the weather over which people will have no influence; all they can do is hope for improved circumstances. Finally, the primary concern is of the utmost importance to shield the crop from diseases of different kinds since the diseases in question might affect the crop's total growth and yield (Khalifa *et al*., 2021). If one can, then if these diseases can be identified on time, the crop will then be able to be safeguarded with the proper nutrients. Providing a method for recognizing and categorizing illnesses that are amenable to digitalization would be beneficial for agriculturists. It will shorten the time needed to identify diseases and improve the accuracy with which they are classified (Mahum *et al*., 2023). The authors compared two CNN approaches for diagnosing 26 diseases in 14 crops using the (Hou *et al*., 2021) Plant Village Dataset.

The main contributions of this study are: (1) Histogram Equalization is used for image enhancement; (2) Deep CNN model is used for feature extraction; (3) Feature Concatenation and Wrapper Feature Selection are used for selecting core features; (4) SVM classifier and its Variants are used for Classification.

## Materials and Methods

In this proposed four phases, the first is preprocessing for image enhancements. In preprocessing, we resize the image to 300'300'3. Histogram equalization is utilized for image enhancement. In the second phase, we used feature engineering using Deep CNN models. Darknet-53 (Zhou *et al*., 2014; Deng *et al*., 2009), AlexNet (Zhou *et al*., 2014; Saleem *et al*., 2025), and Vgg-19 (Redmon, 2021) are used for feature extraction. Moreover, in the third phase, feature concatenation is performed to concatenate the in-depth CNN features. Equilibrium Optimization (EO) (Krizhevsky *et al*., 2017) is performed for feature selection. These selected features are given to different SVM (Simonyan & Zisserman, 2014) for Classification. Figure (1) displays a block representation of the Proposed methodology.

### *Preprocessing*

Our proposed method begins with the preprocessing step of resizing (Too and Mirjalili, 2021) and histogram equalization. The resizing method displays each matrix slice as a separate, unconnected pixel. The red pixel intensity is represented on the first larger plane, the green pixel intensity on the second level, and the blue pixel intensity on the third level in a true color image. With this method, we can demonstrate two behaviors in the column while maintaining the same image quality regardless of the column width. It is important to note that only the first two dimensions of the supplied image must be modified. Due to its ease of use and high level of performance, Histogram Equalization (HE).

Keerthi *et al*. (2001) has become one of the most used algorithms for contrast enhancement. As a rule, the HE creates a better picture with a linear cumulative histogram because of the even distribution of pixel values. Histogram modification is often used in conjunction with HE enhancement for applications (Wang *et al*., 2008), including medical image processing, object recognition, and texture. Histogram Equalization (HE) is a technique for improving image contrast that is wholly discussed in Eq. (1):

$$P(v_i) = \frac{n'i}{N''} i = 0, \ldots, L \quad (1)$$

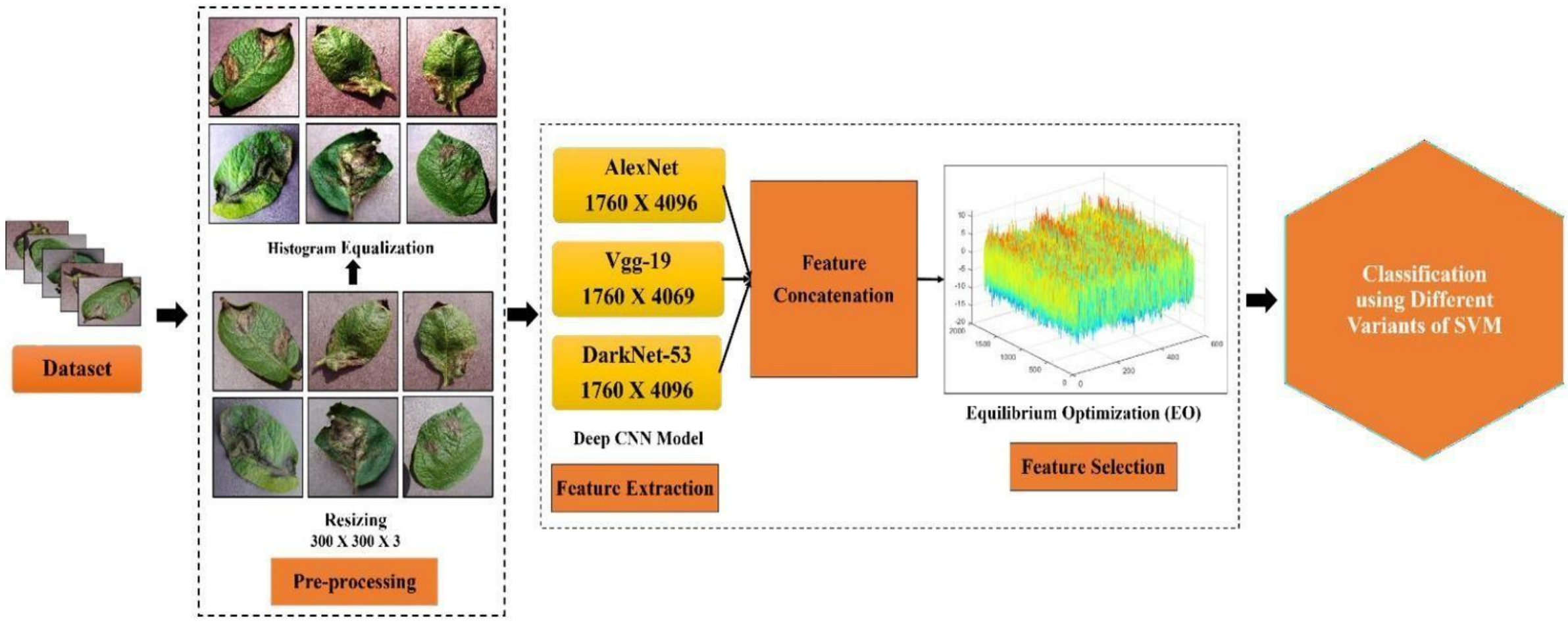

**Fig. 1:** The block diagram of the proposed methodology

In this case, a digital image with grayscale values between [0, L-1] and examine the probability distribution of those values. Eq (1) is utilized to calculate the image's function. If $n'i$ is the number of pixels in the image that have the grey level $vi$ and $vi$ is the equivalent grey level. The CDF may also be calculated in the following ways:

$$C(v_i) = \sum_{j=0}^{j=i} P(r), i = 0, L-1, 0 \leq C(v_i) \leq 1 \quad (2)$$

To rectify this, Histogram Equalization (HE) uses the following equation to adjust the input image's grayscale from level $y_i$ to level $v_i$ (2). As a result:

$$y_i = (L-1) X\, C(v_i) \quad (3)$$

$$\Delta y_i = (L-1) X\, P(v_i) \quad (4)$$

Gray level $\Delta y_i$ Changes can be computed in the usual histogram equalization method as shown in the above Eqs. (3-4). Figure (2) displays the results of the histogram equalization of late blight leaf disease.

### *Feature Engineering*

In this phase, we implemented feature engineering using a deep neural network. Feature extraction is accomplished with the help of Darknet-53, AlexNet, and Vgg-19. The Global Average Pooling (GAP) layer is used in the darknet for feature extraction. The 1760'1024 are total features extracted using the GAP layer of Darknet53. Moreover, the FC7 layer of AlexNet and Vgg-19 is used for feature extraction. A total of 4096 features are selected individually and extracted. Following the completion of feature extraction, the next step is feature concatenation, which is used to combine the deep CNN features. It is mathematically discussed in Eq (5). The 1760'9216 are total features after feature concatenation as described in Eq. (6). For feature optimization, Equilibrium Optimization (EO) is often carried out. These core selected features are sent to several SVM variants for classification. Figure (3) shows the mesh representation of feature engineering:

$$F_c = \sum_{N=1}^{1024} Darknet53 \cup \sum_{N=1}^{4096} AlexNet \cup \sum_{N=1}^{4096} V.gg19 \quad (5)$$

$$F_c = \sum_{N=1}^{9216} Feat(TotalFeatures) \quad (6)$$

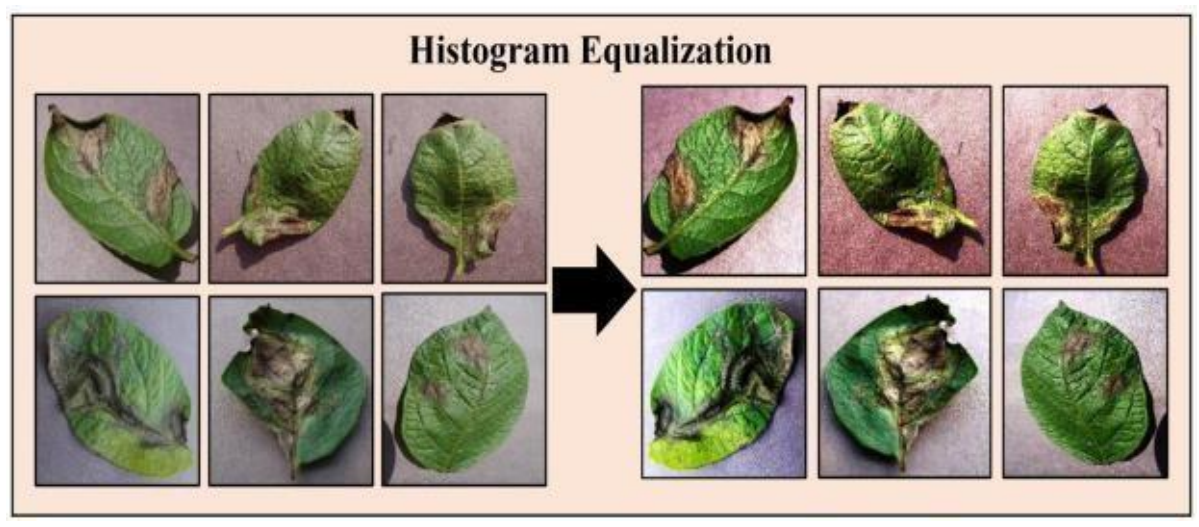

**Fig. 2:** Result of histogram equalization on leaf late blight images

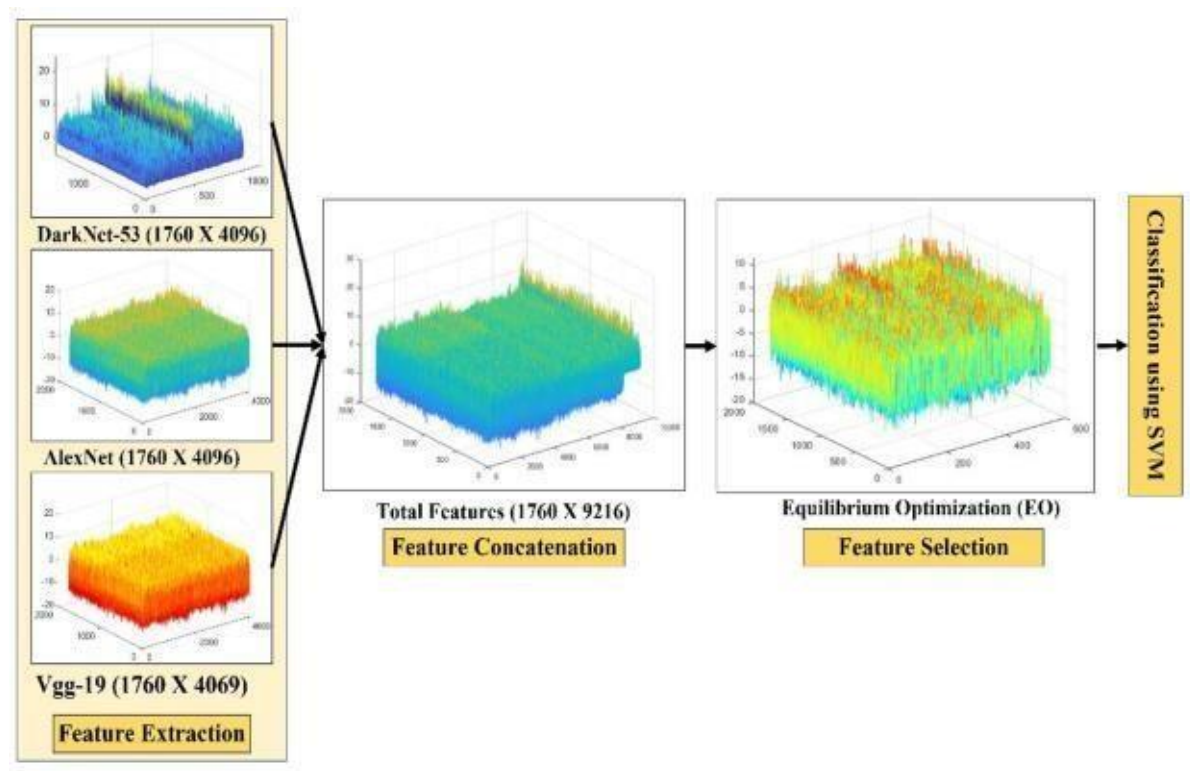

**Fig. 3:** Mesh representation of feature engineering

## Results

Detailed experiments are carried out in this part to test the proposed framework. The following procedures are used to achieve the results: An image is enhanced using histogram equalization; a pre-trained deep CNN model is used for feature extraction; the best deep model features are improved using Equilibrium optimization, and the best features are concatenated for Classification using SVM variants. In this study, the plant village dataset is used. Moreover, only binary Classification (potato healthy and late potato blight, as shown in Fig. 4) is performed using deep learning. Accuracy, Total cost, Prediction speed, Training speed, Precision, F1 Rate, and Recall rate are all computed to evaluate the proposed method's performance. MATLAB is used for all the simulations. Classifiers and the best outcomes in each test case are compared, along with a confusion matrix. Five folds are used in the experiments.

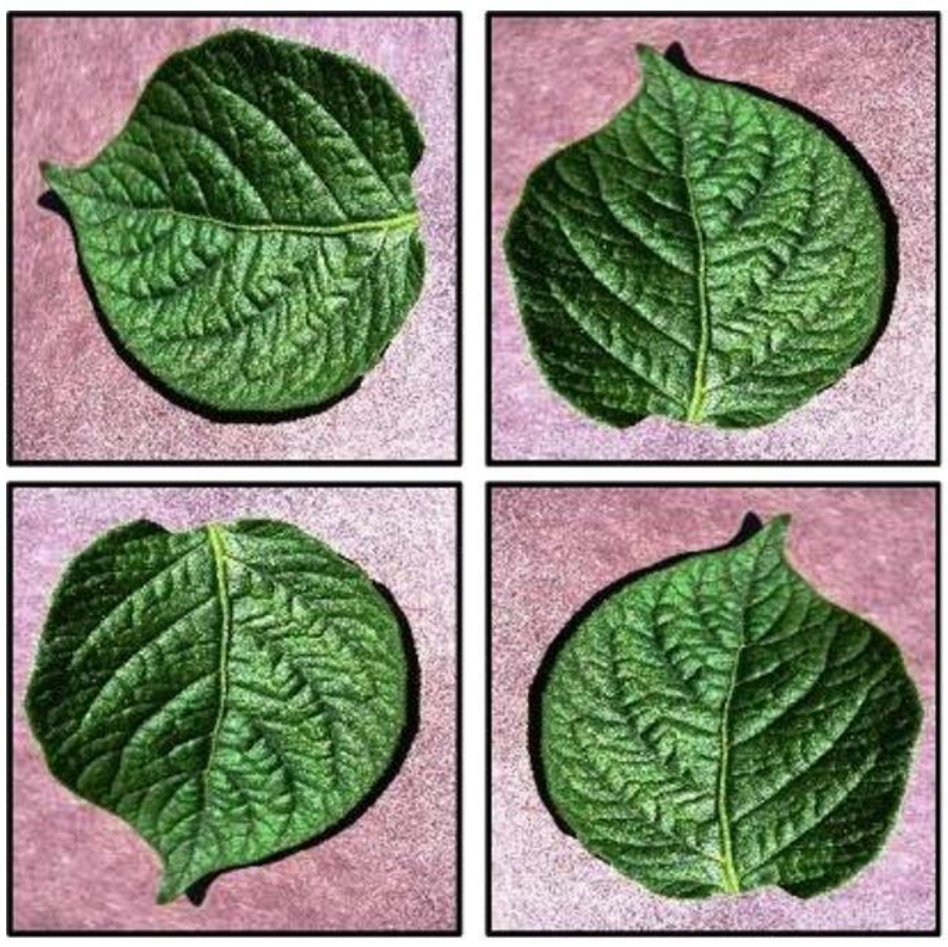

(a)

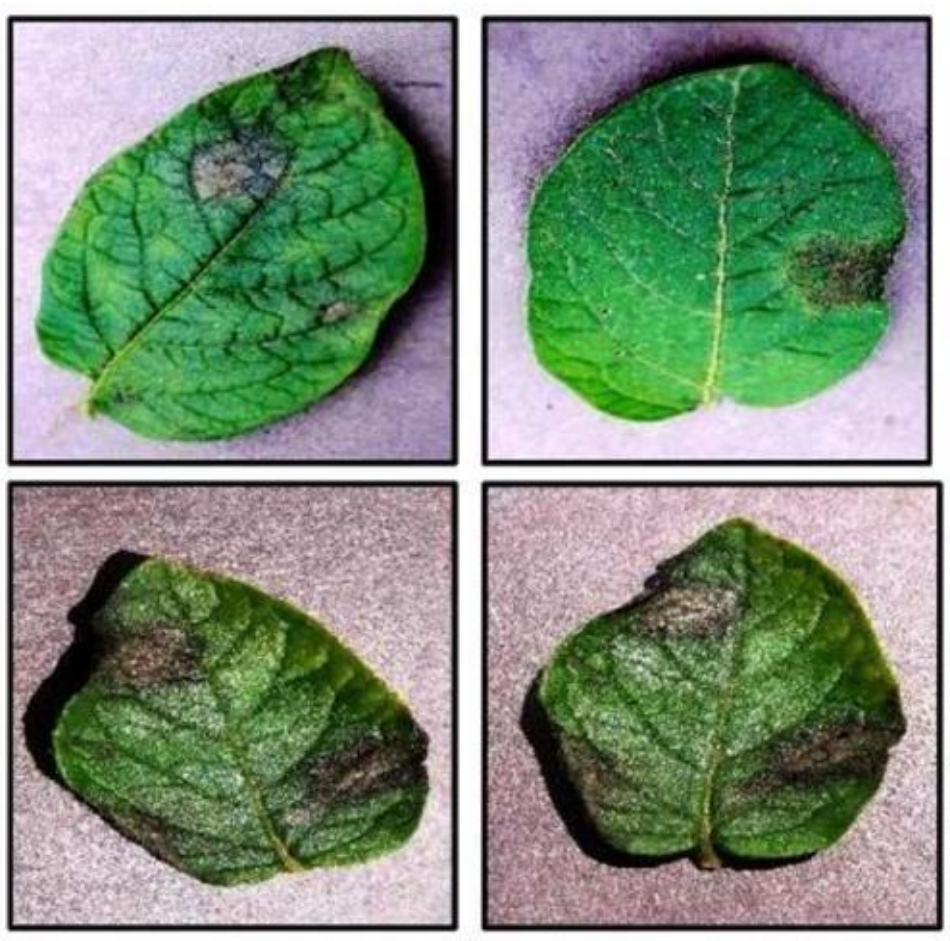

(b)

**Fig. 4:** (a) Healthy potato leaves images; (b) Late blight potato leaves images

Various performance assessment criteria are used in this study to assess the efficacy of the proposed research. Most protocols rely on the confusion matrix created during the identification task's testing process. These protocols are simply calculated as Accuracy (Ac), $Sensitivity$ ($Se$) & $Specificity$ ($Sp$):

$$Ac = \frac{TruePosValues+TrueNegValues}{TruePosValues+TrueNegValues+FalsePosValues+FalseNegValues} \quad (7)$$

$$Se = \frac{TruePosValues}{TruePosValues++FalseNegValues} \quad (8)$$

$$Sp = \frac{TrueNegValues}{TrueNegValues++FalsePosValues} \quad (9)$$

### *Experiment Setup 1 (150 Features)*

This experiment tests a new technique on a realistic sample of 150 features. The chosen feature vector is 1760'150 in dimension. SVM classifiers classify late blight and normal instances based on the feature set. With a score of 98.1 percent, the C-SVM & Q-SVM classifier was the best in this test. L-SVM classifier has the second-highest accuracy. A summary of this experiment's performance using 150 characteristics is shown in Table (1). The graph below shows true classifiers and their respective training times and prediction speeds in Figs. (6-7). Figure (5) displayed confusion matrices for the classifier.

**Table 1:** Experiment Setup 1

| Classifier | Features | Accuracy | Sensitivity | Specificity |
|---|---|---|---|---|
| Linear SVM | 150 | 97 | 98.66 | 94.92 |
| Quadratic SVM | 150 | 98.1 | 99.09 | 96.78 |
| Cubic SVM | 150 | 98.1 | 98.69 | 96.26 |
| Fine Gaussian SVM | 150 | 56.8 | 56.82 | 0 |
| Medium Gaussian SVM | 150 | 97.6 | 99.08 | 95.79 |
| Coarse Gaussian SVM | 150 | 95.6 | 98.53 | 54.01 |

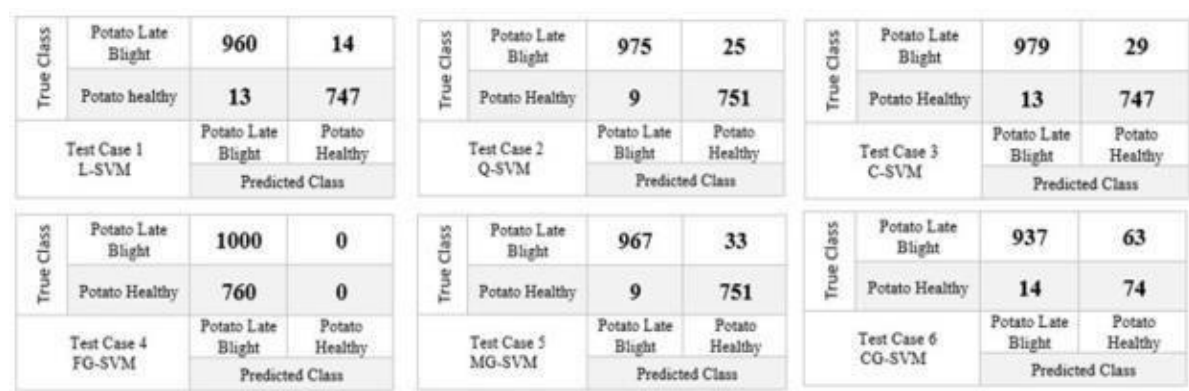

**Fig. 5:** Displayed confusion matrices (150 Features) with respect to classifier

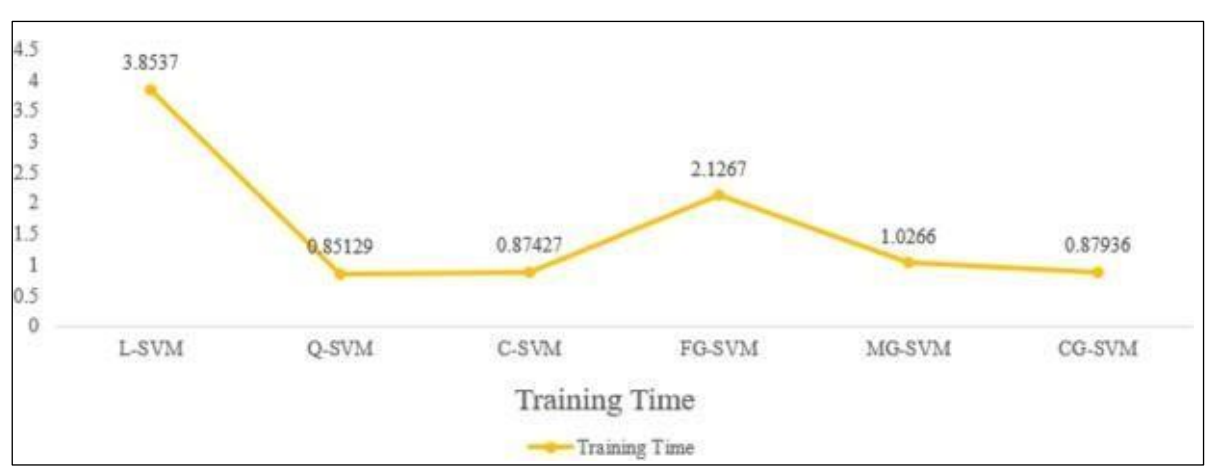

**Fig. 6:** The training time of 150 Features with respect to classifier

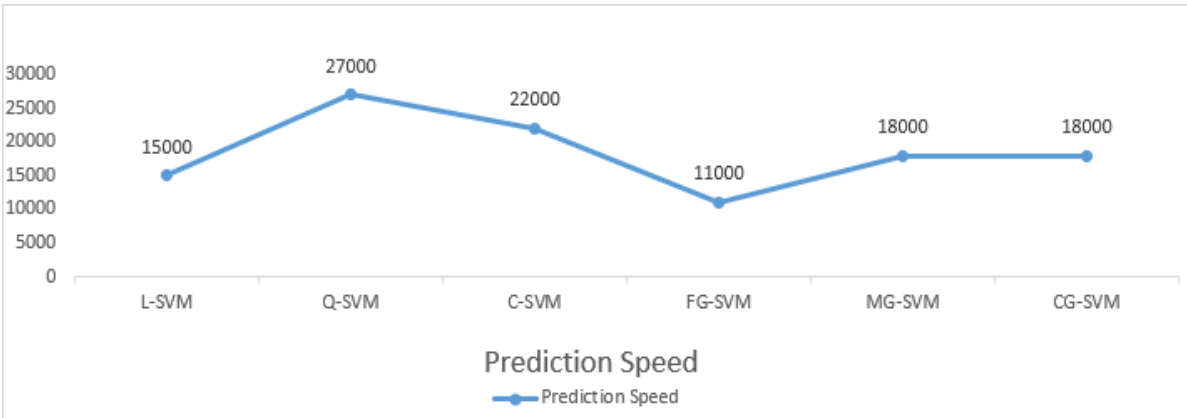

**Fig. 7:** Prediction speed of 150 Features with respect to classifier

**Table 2:** Experiment Setup 2

| Classifier | Features | Accuracy | Sensitivity | Specificity |
|---|---|---|---|---|
| Linear SVM | 250 | 97.6 | 98.68 | 96.14 |
| Quadratic SVM | 250 | 98.5 | 99.29 | 97.54 |
| Cubic SVM | 250 | 98.9 | 99.6 | 98.04 |
| Fine Gaussian SVM | 250 | 56.8 | 56.82 | 0 |
| Medium Gaussian SVM | 250 | 98.4 | 99.39 | 97.04 |
| Coarse Gaussian SVM | 250 | 96.3 | 98.85 | 93.28 |

## *Experiment Setup 2 (250 Features)*

An innovative approach is tested through its trials in this experiment by being applied to a representative sample of 250 features. The dimension chosen for the selected feature vector is 1760 by 250. SVM classifiers differentiate between late blight occurrences and normally based on the feature set. The C-SVM classifier performed the best in this evaluation, obtaining a score of 98.9% overall. The accuracy of the Q-SVM classifier is the second-best in the industry. Table (2) provides a summary of the performance of this experiment based on the evaluation of 250 criteria. A genuine classifier is shown in the graph below, along with their training times and prediction speeds, as found in Figs. (9-10). The confusion matrices with regard to the classifier are given in Fig. (8).

## *Experiment Setup 3 (550 Features)*

This experiment set up a new approach through its tests by applying it to 550 features chosen at random. The selected feature vector has dimensions of 1760 by 550. SVM classifiers are used in the feature set to distinguish late blight outbreaks from normal events. The C-SVM classifier came out on top in this test, with an impressive total accuracy of 99.6%. The Q-SVM classifier has the second-best Accuracy in the business. The results of this trial, as measured by 550 criteria, are summarized in Table (3). Figures (12-13) demonstrate the training and prediction times for one such classifier, respectively. As shown in Fig. (11), the confusion matrices for the classifier were provided.

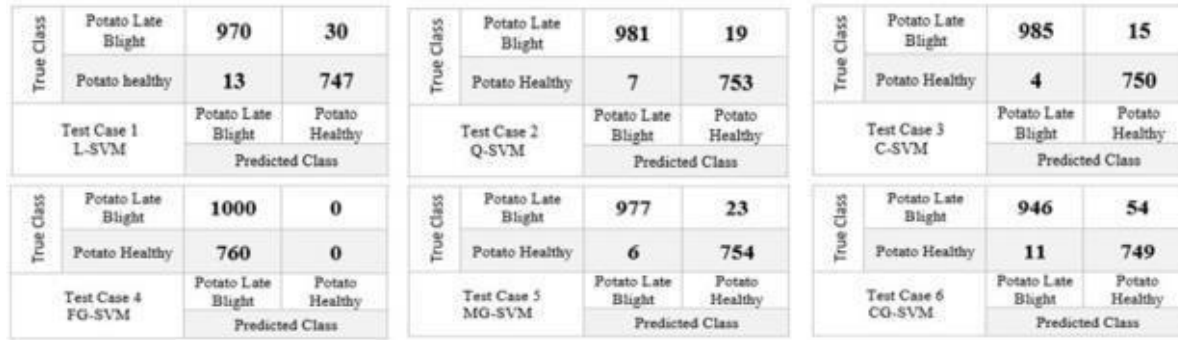

**Fig. 8:** Displayed confusion matrices (250 Features) with respect to classifier

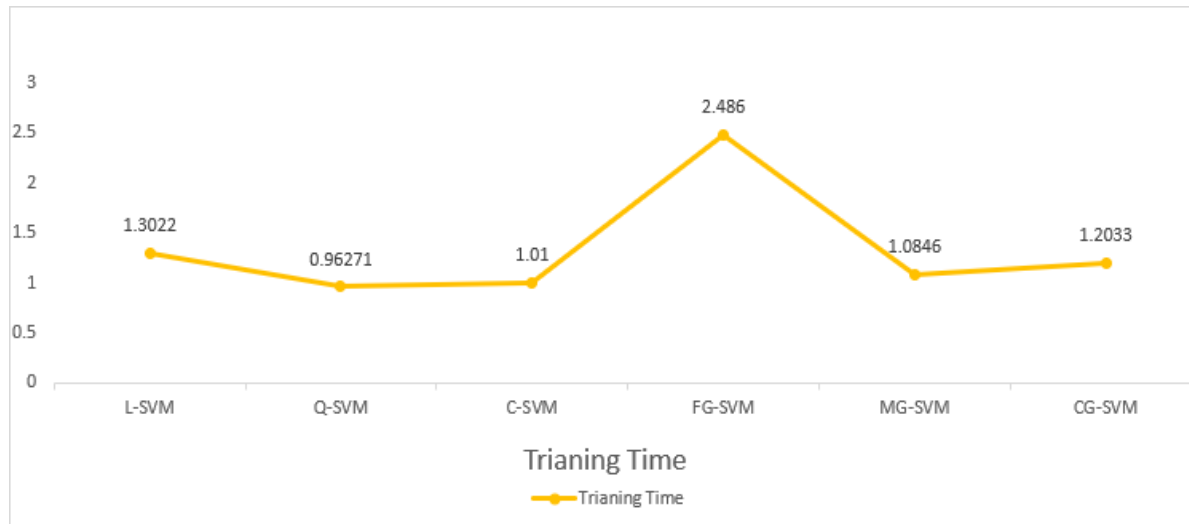

**Fig. 9:** The trainingtime of 250 Features with respect to classifier

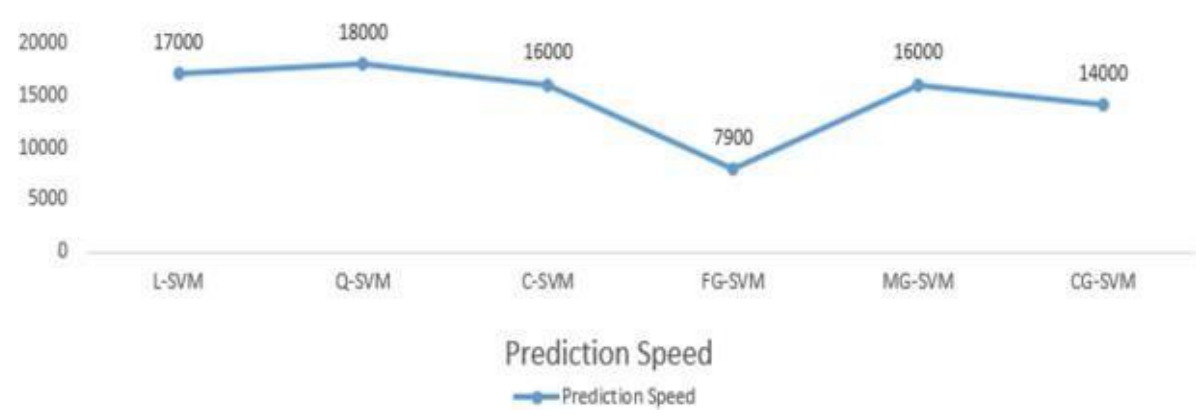

**Fig. 10:** Prediction speed of 250 Features with respect to classifier

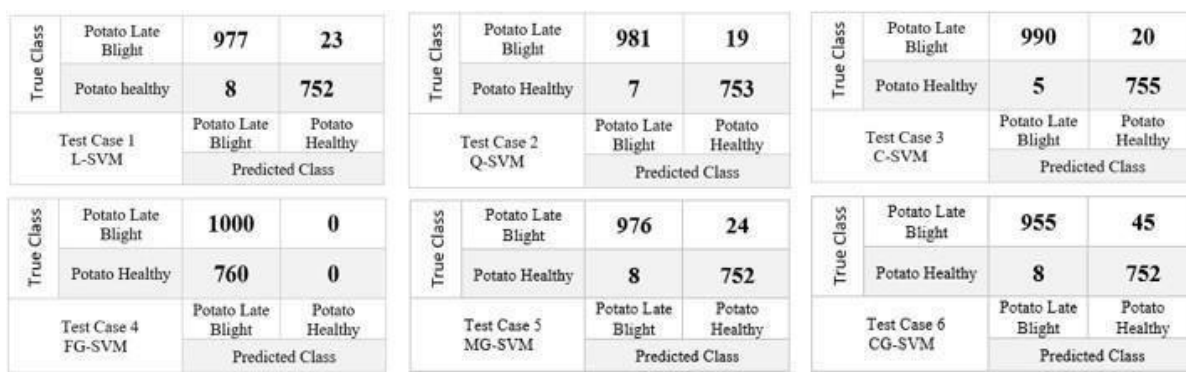

**Fig. 11:** Displayed confusion matrices (550 Features) with respect to classifier

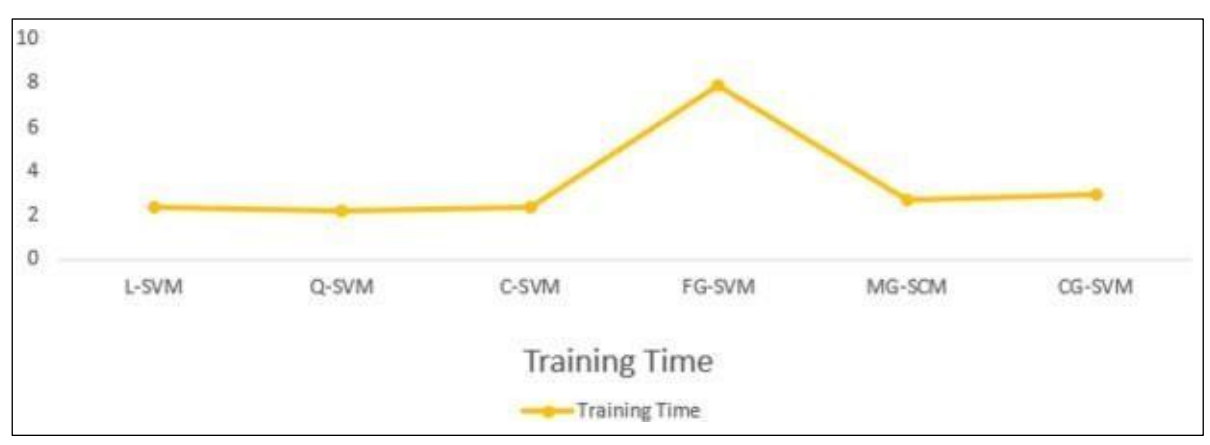

**Fig. 12:** The training time of 550 Features with respect to classifier

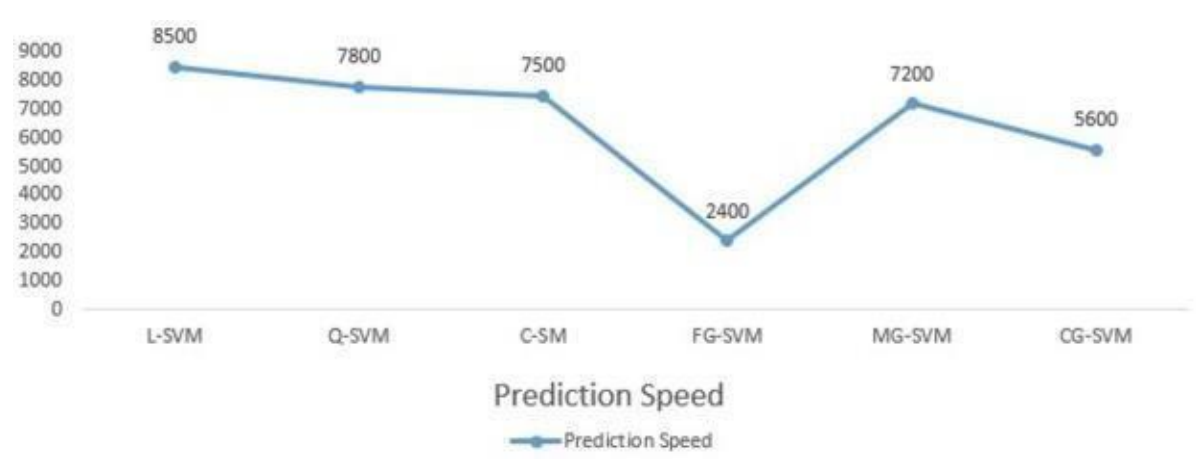

**Fig. 13:** Prediction speed of 550 Features with respect to classifier

**Table 3:** Experiment Setup 3

| Classifier | Features | Accuracy | Sensitivity | Specificity |
|---|---|---|---|---|
| Linear SVM | 550 | 98.2 | 99.19 | 97.03 |
| Quadratic SVM | 550 | 98.5 | 99.29 | 97.54 |
| Cubic SVM | 550 | 99.6 | 99.5 | 97.42 |
| Fine Gaussian SVM | 550 | 56.8 | 56.82 | 0 |
| Medium Gaussian SVM | 550 | 98.2 | 99.19 | 96.91 |
| Coarse Gaussian SVM | 550 | 97 | 99.17 | 93.28 |

## Discussion

This study evaluates a novel method using a substantial data set of 150 features. Dimensions of the selected feature vector are 1760 by 150. SVM classifiers employ the feature set to separate late blight from normal occurrences. The C-SVM & Q-SVM classifier performed the best in this evaluation with a success rate of 98.1%. The Accuracy of the L-SVM classifier is second best. Table (1) provides a brief overview of the results of this experiment based on a set of 150 features. In setup 2 of the experiment, a novel method is put through its paces by being applied to 250 features that are meant to be indicative of the whole. The selected feature vector will have dimensions of 1760 by 250. SVM classifiers are used in the feature set to distinguish late blight occurrences from typical occurrences. The C-SVM classifier came out on top in this test, with an impressive total accuracy of 98.9%. The Q-SVM classifier has the second-best accuracy in the business. The results of this trial were measured using 250 criteria. Using 550 randomly selected characteristics, prepare a new technique for testing in experiment setup 3. The size of the chosen feature vector is 1760 by 550. SVM classifiers are used in the feature set to differentiate late blight outbreaks from regular occurrences. In this evaluation, the C-SVM classifier performed the best, with the highest accuracy of 99.6 percent. Q-SVM is a classifier that has the second-best accuracy in the market. Table (3) provides a concise summary of the 550 criteria-based trial findings.

## Conclusion

In many parts of the globe, the potato is one of the most common staple crops. The cultivation of potatoes has skyrocketed in popularity in recent decades. Several diseases might impede potato growth. The leaves on this plant seem to be severely infected. Early Blight (EB) and Late Blight (LB) are two common leaf diseases that attack potato plants. A significant increase in this crop's productivity might be achieved with the early diagnosis of these diseases. As a result, the application of image processing for diagnosing and evaluating these conditions is very desirable. Here, we describe a self-sufficient approach using image processing and machine learning to identify potato leaf late blight. The beginning of the pipeline's four stages, preprocessing, focuses on enhancing images. During the first processing stage, the image is shrunk to 300'300'3 image enhancement using histogram equalization. Second, we implemented feature engineering using Deep CNN models. Features are extracted using Darknet-53, AlexNet, and Vgg-19. In addition, the deep CNN features are concatenated in the third step. Feature selection is an Equilibrium Optimization (EO) task. Then, we use many different variants of Support Vector Machines (SVMs) to classify based on these features.

## Acknowledgment

We sincerely thank the publisher for their support in publishing this research article. We are grateful for the resources and platform provided, which allowed us to share our findings with a broader audience. Your efforts are greatly appreciated.

## Funding Information

The authors have not received any financial support or funding to report.

## Author's Contributions

**Muhammad Ahtsam Naeem:** Conceptualization, methodology, software design.

**Muhammad Asim Saleem:** Data curation, written-original draft preparation, interim review, and editing.

**Muhammad Imran Sharif:** Review, and editing.

**Shahzad Akber:** Review, and editing. Visualization, investigation.

**Sajjad Saleem:** Investigation, written-reviewed.

**Zahid Akhtar And Kamran Siddique:** Supervision and written-reviewed and finalization.

## Ethics

This research does not involve any human participants, personal data, or ethical concerns related to privacy, bias, or harm. This work is not submitted or published anywhere else.